\documentclass{article} 
\usepackage{ijcai20,times}

\usepackage{amsmath,amsfonts,bm}

\def\eqref#1{equation~\ref{#1}}

\def\1{\bm{1}}

\DeclareMathAlphabet{\mathsfit}{\encodingdefault}{\sfdefault}{m}{sl}
\SetMathAlphabet{\mathsfit}{bold}{\encodingdefault}{\sfdefault}{bx}{n}

\usepackage{tikz}
\usetikzlibrary{shapes,arrows,positioning}
\usepackage{hyperref}
\usepackage{url}
\usepackage{graphicx,subcaption}
\usepackage{changepage}
\usepackage[capitalise]{cleveref}
\usepackage{natbib}
\setcitestyle{square}

\newcommand{\cmnt}[1]{\ignorespaces}

\tikzstyle{decision} = [diamond, draw, fill=blue!20, 
    text width=4.5em, text badly centered, node distance=3cm, inner sep=0pt]
\tikzstyle{block} = [rectangle, draw, fill=blue!20, 
    text width=6em, text centered, rounded corners, minimum height=4em,node distance =4cm]
\tikzstyle{line} = [draw, -latex']

\title{\centering Language is Power: Representing States Using \\ Natural Language in Reinforcement Learning}

\author{Erez Schwartz, Guy Tennenholtz, Chen Tessler and Shie Mannor}

\begin{document}

\maketitle

\begin{abstract}
    Recent advances in reinforcement learning have shown its potential to tackle complex real-life tasks. However, as the dimensionality of the task increases, reinforcement learning methods tend to struggle. To overcome this, we explore methods for representing the semantic information embedded in the state. While previous methods focused on information in its raw form (e.g., raw visual input), we propose to represent the state using natural language. Language can represent complex scenarios and concepts, making it a favorable candidate for representation. Empirical evidence, within the domain of ViZDoom, suggests that natural language based agents are more robust, converge faster and perform better than vision based agents, showing the benefit of using natural language representations for reinforcement learning.
    
\end{abstract}

\section{Introduction}

\begin{quote}
    \emph{``Language is power, in ways more literal than most people think." -- Julia Penelope}
\end{quote}

Deep learning based algorithms use neural networks in order to learn feature representations that are good for solving high dimensional machine learning (ML) tasks. Reinforcement learning (RL) is a subfield of ML that has been greatly affected by the use of deep neural networks as universal function approximators \citep{csaji2001approximation}.
These deep neural networks are used in RL to estimate value functions, state-action value functions, policy mappings, next-state predictions, rewards, and more \citep{mnih2015human,schulman2017proximal} using their representation power,
thus combating the ``curse of dimensionality" \citep{powell2007approximate}.

The term representation is used differently in different contexts. For the purpose of this paper we define a \textbf{semantic representation} of a state as one that reflects its meaning as it is understood by an expert. The semantic representation of a state should thus be paired with a reliable and computationally efficient method for extracting information from it. Previous success in RL has mainly focused on representing the state in its raw form  (e.g., visual input in Atari-based games \citep{mnih2015human}). This approach stems from the belief that neural networks (specifically convolutional networks) can extract meaningful features from complex inputs. In this work, we challenge current representation techniques and suggest to represent the state using natural language, similar to the way we, as humans, summarize and transfer information efficiently from one to the other \citep{sapir2004language}.

The ability to associate states with natural language sentences that describe them is a hallmark of understanding representations for reinforcement learning. Humans use rich natural language to describe and communicate their visual perceptions, feelings, beliefs, strategies, and more. The semantics inherent to natural language carry knowledge and cues of complex types of content, including events, spatial relations, temporal relations, semantic roles, logical structures, support for inference and entailment, as well as predicates and arguments \citep{abend2017state}. The expressive nature of language can thus act as an alternative semantic state representation. 

Over the past few years, Natural Language Processing (NLP) has shown an acceleration in progress on a wide range of downstream applications ranging from Question Answering \citep{kumar2016ask}
, to Natural Language Inference \citep{parikh2016decomposable}
through Syntactic Parsing \citep{williams2018latent}
. Recent work has shown the ability to learn flexible, hierarchical, contextualized representations, obtaining state-of-the-art results on various natural language processing tasks \citep{devlin2018bert}. A basic observation of our work is that natural language representations are also beneficial for solving problems in which natural language is not the underlying source of input. Moreover, our results indicate that natural language is a strong alternative to current complementary methods for semantic representations of a state.

\begin{figure}[t!]
    \centering
    \includegraphics[width=0.7\linewidth]{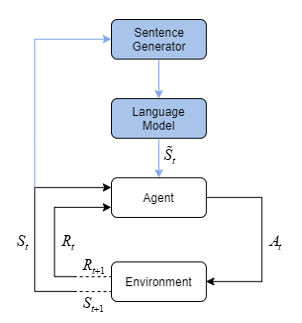}
    \caption{Block diagram for using language-based state representation in RL.}
    \label{fig: block diagram}
\end{figure}

In this work we assume a state can be described using natural language sentences. We use distributional embedding methods\footnote{Distributional methods make use of the hypothesis that words which occur in a similar context tend to have similar meaning \citep{firth1957synopsis}, i.e., the meaning of a word can be inferred from the distribution of words around it. For this reason, these methods are called “distributional” methods.} in order to represent sentences, processed with a standard Convolutional Neural Network for feature extraction. \cref{fig: block diagram} depicts a block-diagram of the basic framework we rely on -- described in detail in Sections~\ref{sec: preliminaries} and~\ref{sec: doom}. We discuss possible semantic representations in Section~\ref{sec: semantic representation methods}, namely, raw visual inputs, semantic segmentation, and natural language representations. Then, in Section~\ref{section: experiments} we compare NLP representations with their alternatives. Our results suggest that representation of the state using natural language can achieve better performance, even on difficult tasks, or tasks in which the description of the state is saturated with task-nuisances \citep{achille2018emergence}. 
Moreover, we observe that NLP representations are more robust to transfer and changes in the environment. We conclude the paper with a short discussion and related work.

\section{Preliminaries}
\label{sec: preliminaries}
\subsection{Reinforcement Learning}

We consider a discounted Markov Decision Process (MDP) \citep{sutton1998reinforcement}. An MDP is defined as the tuple $(\mathcal{S}, \mathcal{P}, \mathcal{A}, r, \gamma)$, where $\mathcal{S}$ is the state space, $\mathcal{A}$ is the action space, $\mathcal{P} : \mathcal{S} \times \mathcal{S} \times \mathcal{A} \rightarrow [0, 1]$ is a transition kernel, $r$ is the reward function, and $\gamma \in [0, 1)$ is the discount factor. 

The goal in RL is to learn a policy $\pi(s)$ which maximizes the reward-to-go $\sum_{t=0}^\infty \gamma^t r_t$. For a given policy, there are two common quantities which are used to estimate the performance: the value function $v^\pi (s) = \mathbb{E}^{\pi} [\sum_t \gamma^t r_t | s_0 = s ]$ and action-value function $Q^\pi (s, a) = \mathbb{E}^{\pi} [\sum_t \gamma^t r_t | s_0 = s, a_0 = a ]$. Intuitively, $v^\pi (s)$ represents how ``good" a state $s$ is and $A^\pi (s,a) = Q^\pi (s,a) - v^\pi (s)$ represents how the agent would improve by playing action $a$ at state $s$ when compared to the current policy $\pi$. Two prominent algorithms for solving RL tasks, which we used in our work, are the value-based DQN \citep{mnih2015human} and the policy-based PPO \citep{schulman2017proximal}.

\paragraph{Deep Q Networks (DQN):}

The DQN algorithm approximates the optimal Q function with a Convolutional Neural Network (CNN), by optimizing the network weights such that the expected Temporal Difference (TD) error of the optimal `bellman Equation is minimized:
\begin{equation*}
    || r(s_t) + \gamma \max_{a' \in \mathcal{A}} Q(s_{t+1}, a'; \theta_\text{target}) - Q(s_t, a_t; \theta) ||_2^2 \, .
\end{equation*}
The DQN is an off-policy learning algorithm, i.e., transitions are stored in the Experience Replay and used for training. When applying minibatch training updates, the DQN samples tuples of experience at random from the pool of stored samples in the buffer. In addition, the DQN maintains two separate Q-networks: the current Q-network with parameters $\theta$, and a target Q-network with parameters $\theta_\text{target}$. Finally, every $T_\text{update}$ (fixed number of iterations) the target network is updated to be equal to the current one.

\paragraph{Proximal Policy Optimization (PPO):}

In contrast to the DQN, PPO is an on-policy algorithm. PPO uses two estimators, one for the policy $\pi$, and one for the value $v^\pi$. Training is performed by interleaving data collection and policy optimization. Once a batch of new transitions is collected, PPO attempts to maximize the objective
\begin{equation*}
    \min \left( \frac{\pi (a| s)}{\pi_\text{old} (a | s)} \hat A^\text{GAE}, \text{clip}\!\left( \frac{\pi (a| s)}{\pi_\text{old} (a | s)}, 1 - \epsilon, 1 + \epsilon \right)\! A^\text{GAE} \right),
\end{equation*}
where $\frac{\pi (a| s)}{\pi_\text{old} (a | s)}$ is the ratio of the probability of taking the given action under the current policy $\pi$ to that of taking the same action under the previous policy $\pi_\text{old}$ used to collect the data. $\epsilon$ is a hyperparameter which controls the amount of clipping. Intuitively, the loss encourages the policy to improve its behavior whereas the clipping prevents it from deviating, thereby forcing conservative updates. 

\section{Semantic Representation Methods}
\label{sec: semantic representation methods}

Contemporary methods for semantic representation of states follow one of three approaches: (1) raw visual inputs \citep{mnih2015human,kempka2016vizdoom}, in which raw sensory values of pixels are used from one or multiple sources, (2) feature vectors \citep{todorov2012mujoco}, in which general features of the problem are chosen, with no specific structure, and (3) semantic segmentation maps \citep{ronneberger2015u}, in which discrete or logical values are used in one or many channels to represent the general features of the state. 

In RL, the raw form often corresponds to pixels representing an image. However, the image is only one form of a semantic representation. In semantic segmentation, the image is converted from a 3-channel (RGB) matrix into an $N$-channel matrix, where $N$ is the number of classes. In this case, each channel represents a class, and a binary value at each coordinate denotes whether or not this class is present in the image at this location. 
Semantic segmentation maps contain less \textbf{task-nuisances} \citep{achille2018emergence} than raw images. Task nuisances are random variables that affect the observed data, but are not informative to the task we are trying to solve.

In this paper we propose an additional method for representing a state, namely using natural language descriptions. One method to achieve such a representation is through image captioning \citep{hossain2019comprehensive}. 
Natural language is both rich as well as flexible. This flexibility enables the algorithm designer to represent the information present in the state as efficiently and compactly as possible. In addition, natural language can contribute to establishing interpretable agents, a key component of Explainable Artificial Intelligence \citep{samek2017explainable}.

\begin{figure}[t!]
\centering
\begin{subfigure}{.4\textwidth}
    \centering
    \includegraphics[width=\textwidth]{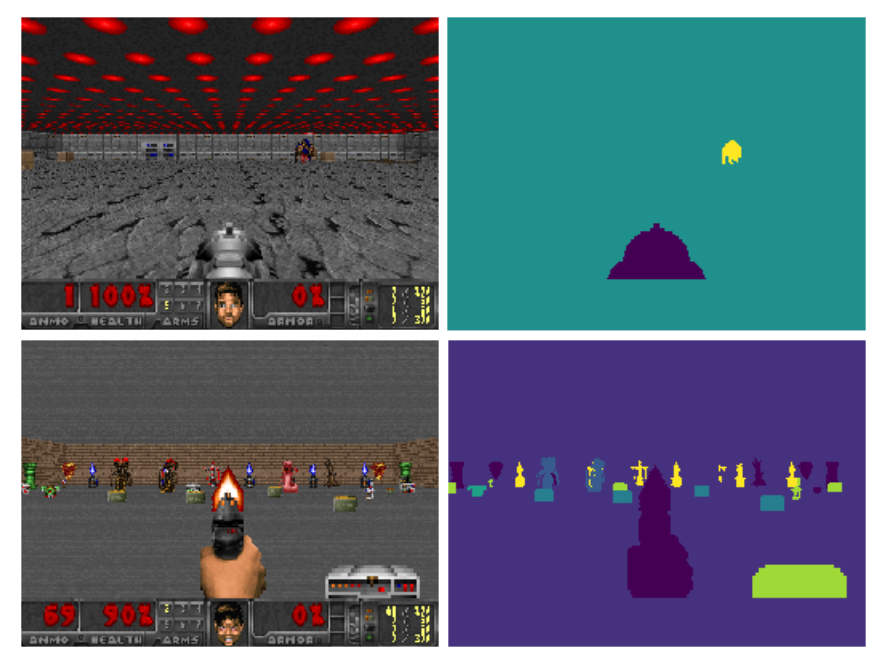}
\end{subfigure}
\caption{
Raw visual inputs and their corresponding semantic segmentation in the VizDoom enviornment. 
}
\label{fig: representations in vizdoom}
\end{figure}

\subsection{Natural Language State Representation}
\label{sec: text cnn}
In this work, we represent the state using natural language, i.e., \emph{words}. One approach for representing natural language sentences is through word embeddings. A word embedding is a mapping from a word $w$ to a vector $\mathbf{w} \in \mathbb{R}^d$. A simple form of word embedding is the Bag of Words (BoW) -- a vector ${\mathbf{w} \in \mathbb{N}^{|D|}}$ ($|D|$ is the dictionary size), in which each word receives a unique 1-hot vector representation. Recently, more efficient methods have been proposed, in which the embedding vector is smaller than the dictionary size, $d \ll |D|$. These methods are also known as distributional embeddings. 

The distributional hypothesis in linguistics is derived from the semantic theory of language usage (i.e., words that are used and occur in the same contexts tend to have similar meanings). Distributional word representations are a fundamental building block for representing natural language sentences. Word embeddings such as Word2vec \citep{mikolov2013efficient} and GloVe \citep{pennington2014glove} build upon the distributional hypothesis, improving efficiency of state-of-the-art language models.

While the common theme in NLP is to use recurrent neural networks, Convolutional Neural Networks (CNNs), originally invented for computer vision, have been shown to achieve strong performance on text classification tasks 
as well as other traditional NLP tasks \citep{collobert2011natural}. In this paper we consider a common architecture, the Text-CNN \citep{kim2014convolutional}, which has been used extensively in NLP domains and enjoys lower complexity and is thus a natural candidate for RL tasks \citep{tessler2019action}.

\section{Natural Language State Representation in the Doom Environment}
\label{sec: doom}

The ViZDoom environment \citep{kempka2016vizdoom} is a 3D world with a relatively realistic physics model and significantly more realistic visual input than Atari~2600 games \citep{mnih2015human}. There, an agent must effectively perceive, interpret, and learn the 3D world in order to make tactical and strategic decisions of where to go and how to act. 

There are two types of visual representations that are provided by the environment: (1) \emph{raw visual inputs}, also the most commonly used form, in which the state is represented by an image from a first person view of the agent, and (2) a \emph{semantic segmentation map} based on the positions and labels of all objects and creatures in the vicinity of the agent. An example of such visual representations in VizDoom is presented in \cref{fig: representations in vizdoom}. Note that semantic segmentation maps are a compact form for representing what a designer believes to be useful features of the state. We therefore expect these representations to obtain superior performance for solving tasks in ViZDoom. Moreover, as we will see in Section~\ref{section: experiments}, a natural language augmentation of these representations can achieve higher performance and robustness.

\subsection{Natural Language Generator}

To simulate natural language based captioning of the state, we constructed a semantic natural language parser. To guarantee a fair comparison against visual representations, we constructed descriptions based on two criteria: limiting information and ambiguity.   
\subsubsection{Limiting Information}
In real world applications, natural language can be efficiently used to describe non-spatial attributes that would be hard otherwise. A designer can inject natural language descriptions with subjective or time-dependent information. Some examples include \textit{``The enemy is too close, watch out!"} and \textit{``An enemy is walking towards you quickly."}. While we believe this aspect of natural language is one of its greater merits, for fair comparison, we limited the expressiveness of our parser to the spatial information present in the semantic segmentation maps.

\begin{figure*}[t!]
\centering
\includegraphics[width=0.28\linewidth]{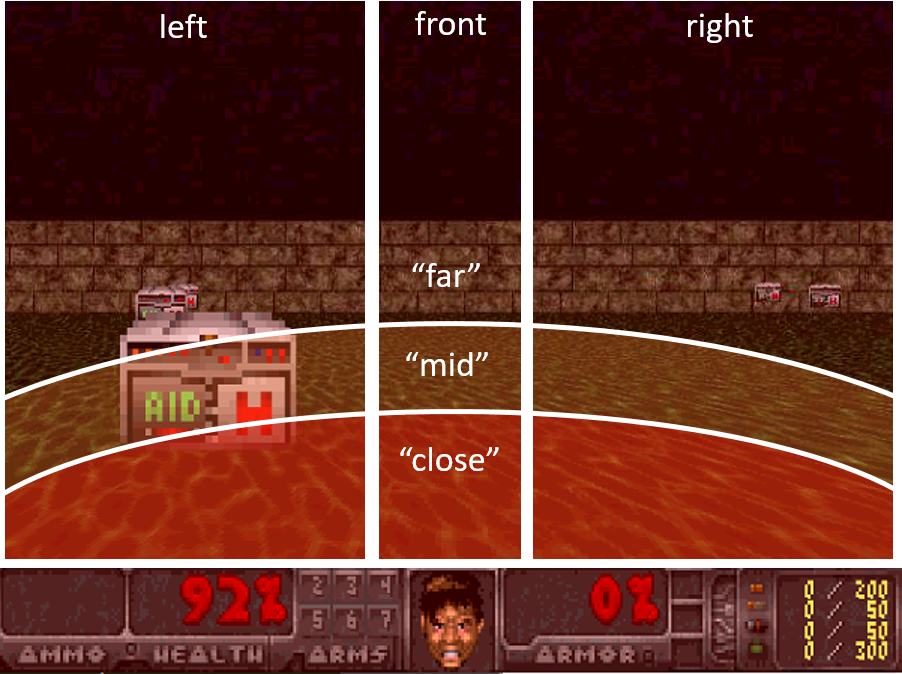}
\hfill
\includegraphics[width=0.28\linewidth]{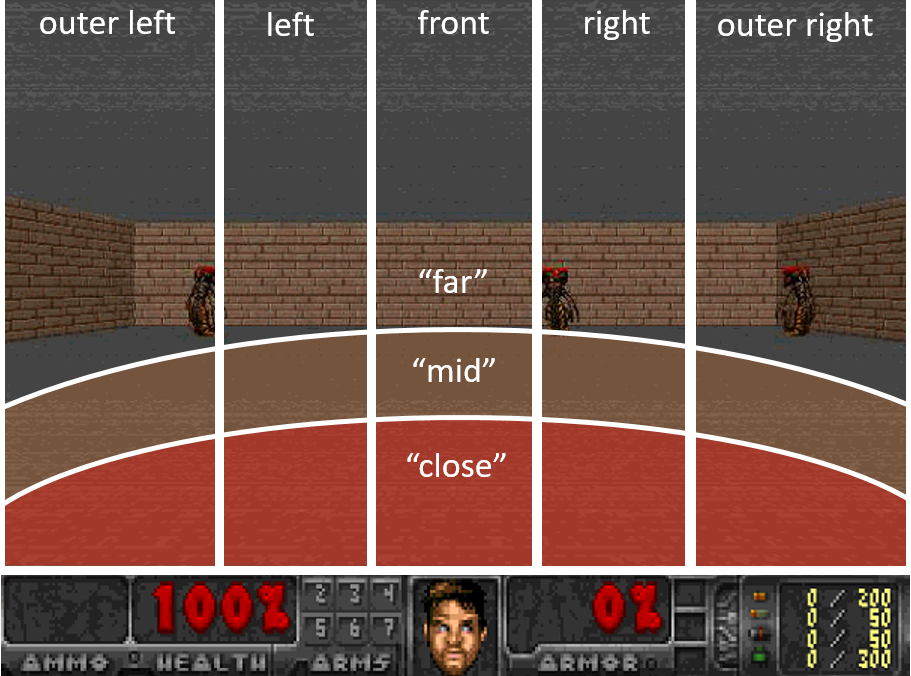}
\hfill
\includegraphics[width=0.28\linewidth]{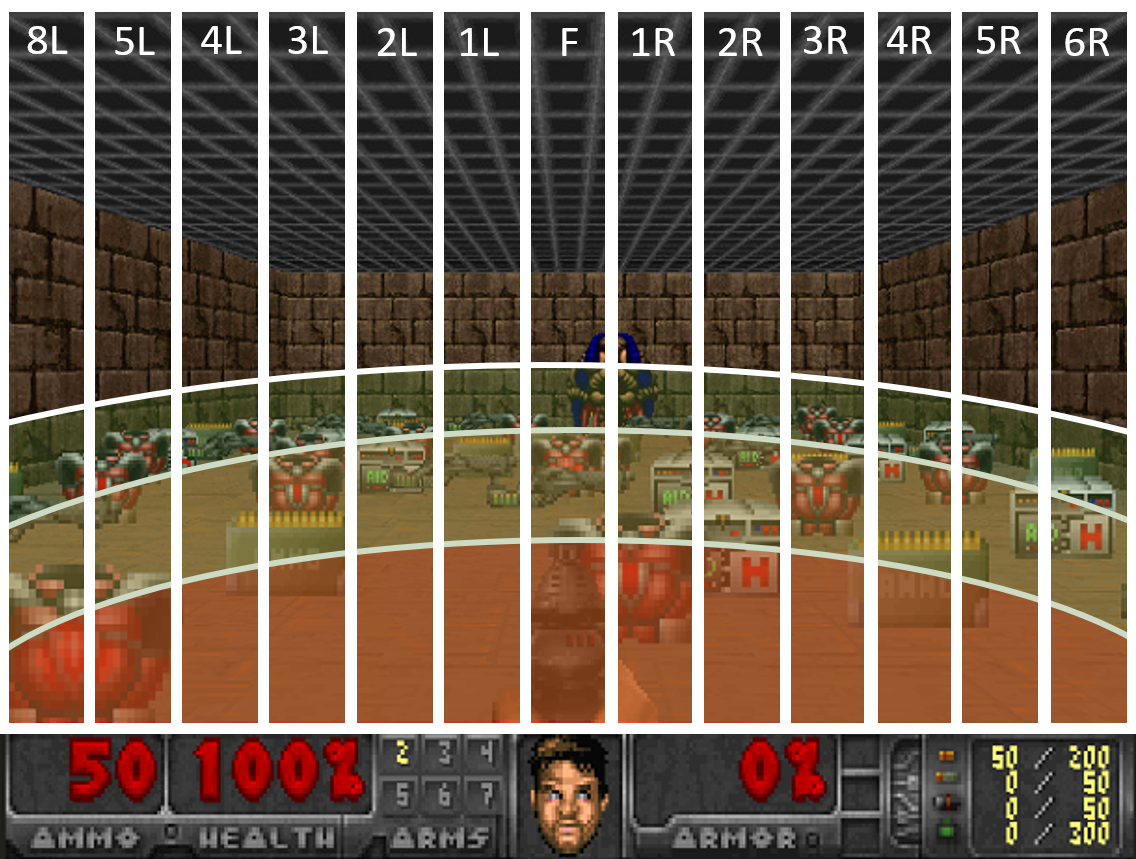}
\caption{Frame division used for describing the state in natural language.}
\label{fig: patches}
\end{figure*}

ViZDoom's semantic segmentation maps were used as the core element for generating our natural language descriptions. Each state of the environment was converted into a sentence based on positions and labels of objects in the frame, as presented in these maps. To implement this, the screen was divided into several vertical and horizontal patches, as depicted in \cref{fig: patches}. These patches describe relational aspects of the state, such as distance of objects and their direction with respect to the agent's point of view. In each patch, objects were counted, and a sentence description was constructed based on this information. Furthermore, to allow for flexibility in the complexity of the state, the number of patches was varied in controlled experiments, as discussed in Section~\ref{sec: discretization}.

\subsubsection{Ambiguity}
A key element of natural language is ambiguity, i.e., there is more than one way to describe a state. For example, all of the following sentences contain information of an agent shooting and missing a target: ``\emph{You shot the wall.}", ``\emph{The player shot and missed.}", and ``\emph{The monster dodged your bullet.}". Coping with ambiguity in natural language is a cornerstone for Explainable Artificial Intelligence \citep{samek2017explainable}, increasing interpretability and transparency of black-box deep learning models. We introduced ambiguity in our parser through a mixture of distinct parsers.

To simulate ambiguity, we constructed ten distinct sentence generators that described each patch of every game frame in a different manner. More specifically, at each step and for every patch, a random parser was sampled, describing the given patch, forming the final state representation. The parsers varied in word usage and sentence structure, allowing for a wide range of descriptions for every state.

\begin{figure*}[t!]
\centering
\begin{subfigure}{\textwidth}
\includegraphics[width=.32\textwidth]{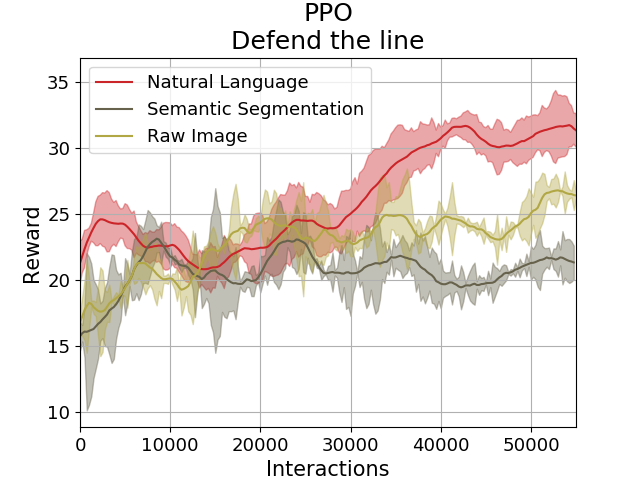}
\includegraphics[width=.32\textwidth]{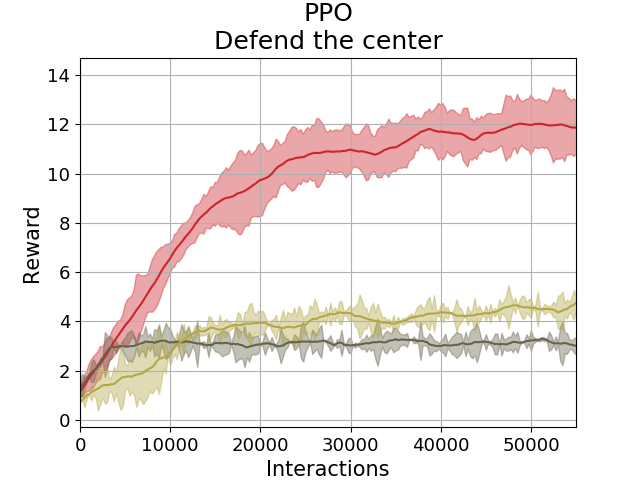}
\includegraphics[width=.32\textwidth]{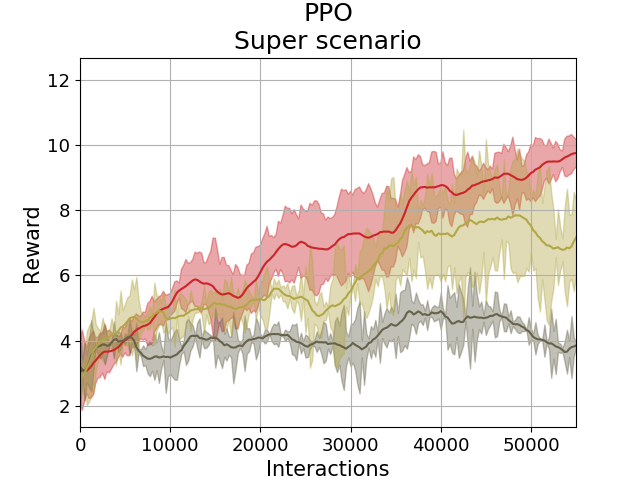}
\end{subfigure}
\hspace{3mm}
\begin{subfigure}{\textwidth}
\includegraphics[width=.32\textwidth]{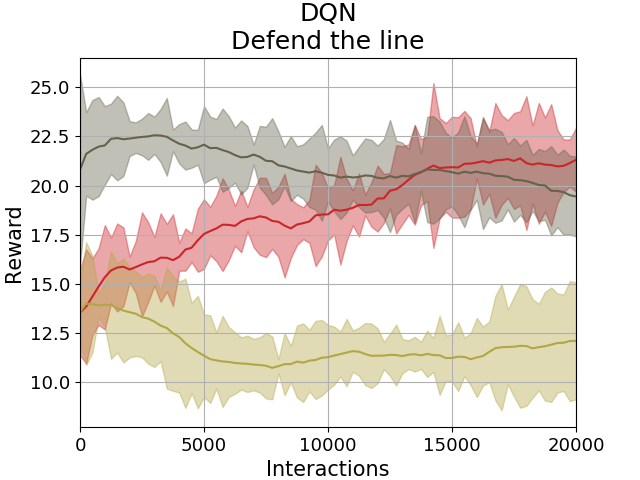}
\includegraphics[width=.32\textwidth]{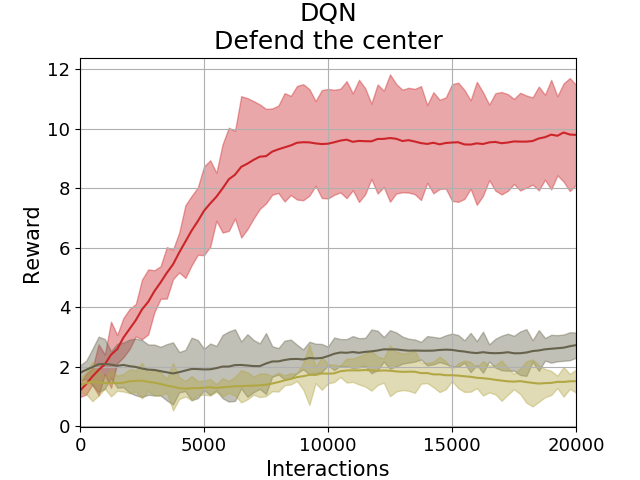}
\includegraphics[width=.32\textwidth]{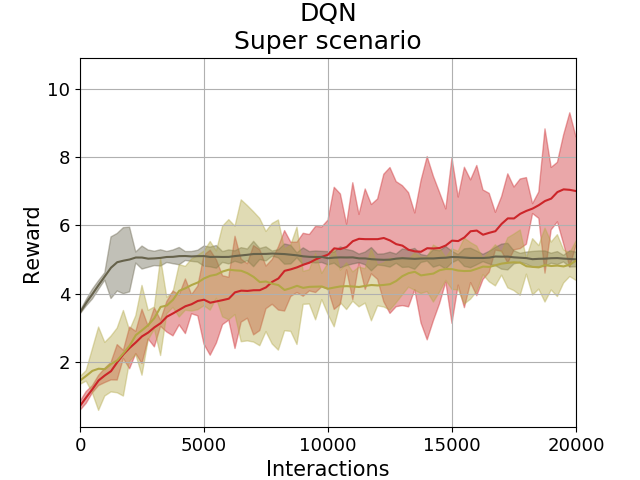}
\end{subfigure}

\caption{Comparison of representation methods on the different VizDoom scenarios using DQN and PPO agents. X and Y axes represent the number of iterations and cumulative reward, respectively.}
\label{fig: scenario comparison}
\end{figure*}

\section{Experiments}
\label{section: experiments}

We tested natural language state representation using our parser against visual-based representations on several tasks, with varying difficulty. In these tasks, the agent can navigate, shoot, and collect items such as weapons and medipacks. The agent obtains a positive reward when it kills the various enemies in each scenario. Frequently, the agent may suffer from constant health degeneration, for instance, due to poison. The different scenarios include a scenario in which the agent must take cover from inbound fireballs (Defend the Center), a scenario in which the agent must defend itself from charging enemies (Defend the Line), and a ``super" scenario, where a mixture of these scenarios was designed to challenge the agent.

Our agent was implemented using a Convolutional Neural Network as described in Section~\ref{sec: text cnn}. The parsed state was converted into embedded representations of fixed length. The sentence length was defined by the longest sentence, whereas, shorter sentences were padded with zeros. In this work, we tested both a DQN and PPO based agent and compared the natural language state representation against the other visual representation techniques; namely, the raw image and semantic segmentation maps.

In order to effectively compare the performance of the different representation methods, we conducted our experiments under similar conditions. First, the same hyper-parameters were used for all representations. Second, to rule out effects of architectural expressiveness, we validated performance over different number of weights and used those that provided best results for visual-based representations. Finally, we ensured the ``super" scenario was positively biased toward image-based representations. This was performed by adding a large amount of items to the game level, thereby filling the state with nuisances. This is especially evident in the NLP representations, as the complexity of learning an optimal policy grows with the sentence's length. This is contrary to image-based representations, in which the input dimension remained constant. We observed an average of over 250 words-per-state.

\subsection{Performance}
Results for DQN and PPO-based agents are presented in \cref{fig: scenario comparison}. Each plot depicts the average reward (across 5 seeds) of all representations methods. It can be seen that NLP representations outperform the visual ones, contrary to the fact that they contain the same information as the semantic segmentation maps. This difference decreases as the complexity of the scenarios increases. Still, NLP representations maintain consistent and robust results throughout all tested scenarios. 

A more thorough analysis of the performance shows that although natural language eventually outperforms other representations, the difference and convergence time are domain dependent. For example, in the Defend the Line scenario, NLP representations did not have a clear advantage over semantic segmentation maps with the DQN agent. 

Additionally, our results indicate that performance of different semantic representations are algorithm-dependent. Interestingly, raw visual inputs showed superiority over semantic segmentation maps when used by the PPO agent, whereas the reverse occurred with the DQN agent. This suggests that the effect of representations on performance is algorithm dependent. We note that this is not related to characteristics of the architecture, but rather to the semantic representation technique and the information required to solve the task.

\subsection{Robustness to Task Nuisances}
\begin{figure}[t!]
\centering
~~~~~PPO ~~~~~~~~~~~~~~~~~~~~~~~~~~~~~~~~~~~~~ DQN \\
\includegraphics[width=0.238\textwidth]{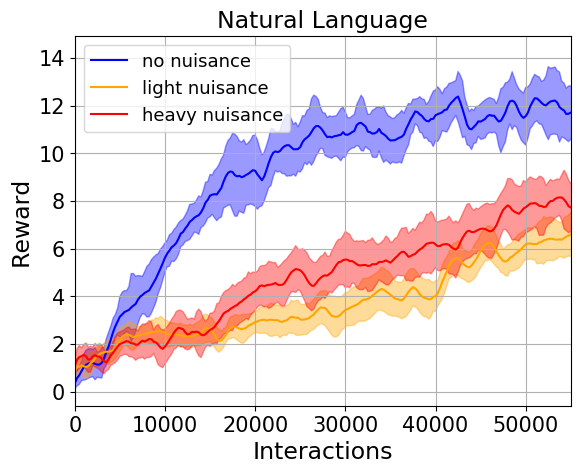}
\includegraphics[width=0.238\textwidth]{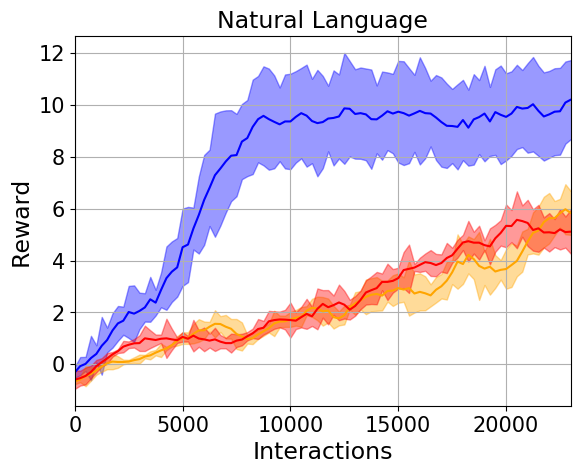} \\
\includegraphics[width=0.238\textwidth]{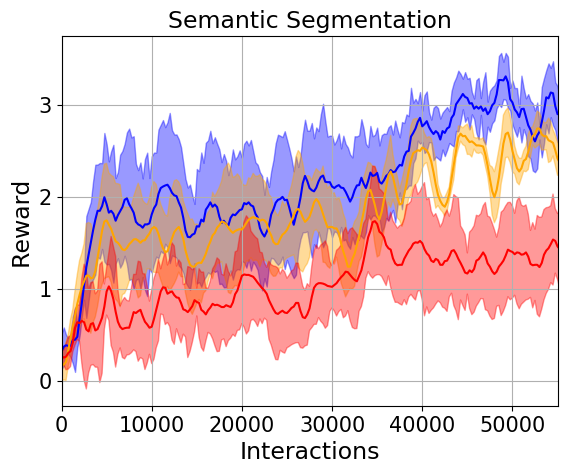}
\includegraphics[width=0.238\textwidth]{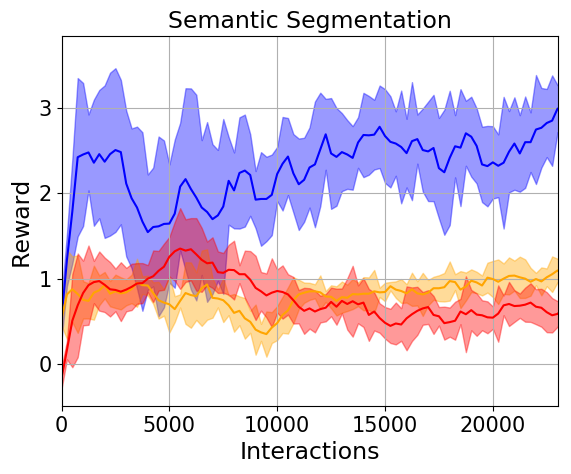} \\
\includegraphics[width=0.238\textwidth]{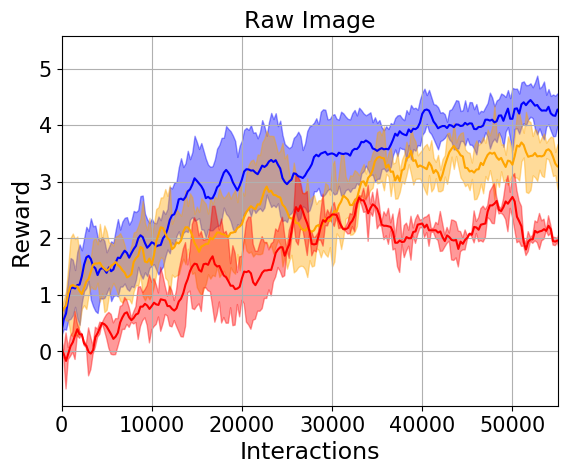}
\includegraphics[width=0.238\textwidth]{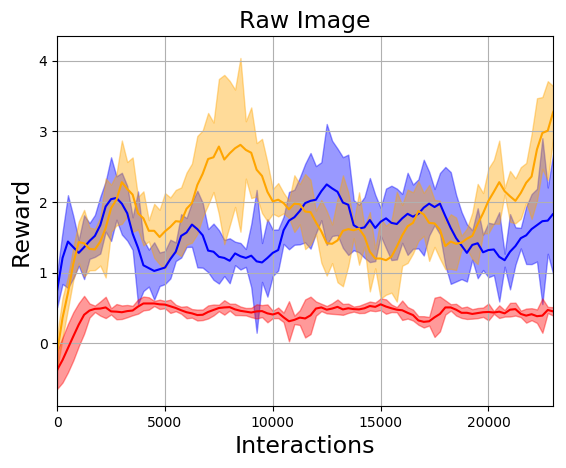}
\caption{ Robustness of each representation type with respect to amount of nuisance, tested on "Defend the Center" scenario.}
\label{fig: nuisance scenarios}
\end{figure}

\cref{fig: nuisance scenarios} shows the effect of task-nuisances over the different representation types. There, a large amount of unnecessary objects were spawned in the level. It can be seen that inflation of the state space with task-nuisances impaired performance of all representations. These increased the state description length to over 250 words, whilst retaining the same amount of useful information. Nevertheless, in most scenarios, NLP representations maintained robustness to the applied noise, outperforming the vision based representations. Moreover, since NLP representations were confronted with ambiguity through random parsers, they were effectively influenced by multiple sources of noise throughout training.

The robustness of NLP representations to task-nuisances indicate their ability to summarize information well. Particularly, the ability to count the elements in each patch mitigated the nuisance effect. We note that, while the raw visual representations were least affected by nuisance, their overall performance remained much smaller than the other representation techniques.

\subsection{Discretization to Patches}
\label{sec: discretization}
Discretization is a commonly used method in RL for lowering the dimensionality of the state space or action space. It allows to efficiently solve problems with a large state space or continuous action space. This approach may come at the expense of finding a suboptimal solution. Nevertheless, adaptive discretization techniques may be used to obtain better overall performance \citep{sinclair2019adaptive}. 

\begin{figure}[t!]
\centering
\begin{subfigure}{.238\textwidth}
\includegraphics[width=\textwidth]{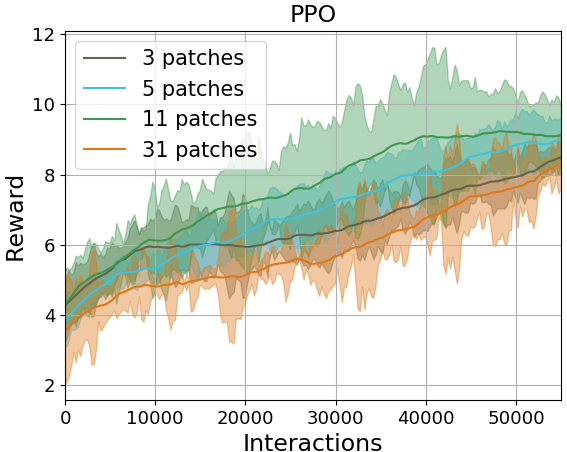}
\end{subfigure}
\begin{subfigure}{.238\textwidth}
\includegraphics[width=\textwidth]{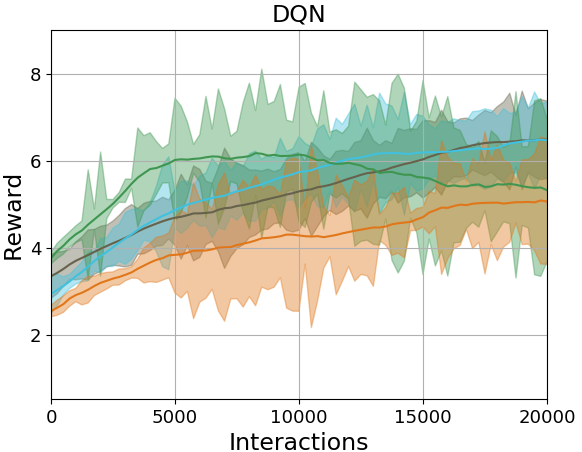}
\end{subfigure}
\caption{Average rewards of NLP based agent as a function of the number of patches in the language model.}
\label{fig: patch count}
\end{figure}

The discretization of the frame to patches in the ViZDoom environment was not carried out for reasons of efficiency, but rather to devise the semantic language parser. Still, it is essential to understand its effect on the performance of the agent. To test the effect of discretization of the frame to patches, we conducted experiments with varying amounts of horizontal patches, ranging from 3 to 31 patches, in the extreme case. Our results, as depicted in \cref{fig: patch count}, suggest that the amount of discretization has a negligible effect on the performance of the NLP-based agents.

These experiments were conducted over the ''super" scenario to ensure a high probability of non-empty patches. The extreme case of 31 patches rendered sentence lengths of over 400 words with no evident loss in performance, indicating that sentence length did not have an effect on performance. Moreover, these results suggest that discretization in the ViZDoom environment does not affect performance.

\section{Related Work}
\label{sec: related work}

Work on representation learning is concerned with finding an appropriate representation of data in order to perform a machine learning task \citep{goodfellow2016deep}. In particular, deep learning exploits this concept by its very nature \citep{mnih2015human}. Work on representation learning include Predictive State Representations (PSR) \citep{littman2002predictive}, which capture the state as a vector of predictions of future outcomes, and a
Heuristic Embedding of Markov Processes (HEMP) \citep{engel2001learning}, which learns to embed transition probabilities using an energy-based optimization problem. 

There has been extensive work attempting to use natural language in RL. Efforts that integrate language in RL develop tools, approaches, and insights that are valuable for improving the generalization and sample efficiency of learning agents. Previous work on language-conditioned RL has considered the use of natural language in the observation and action space. Environments such as Zork and TextWorld \citep{cote2018textworld,tessler2019action} have been the standard benchmarks for testing text-based games. Nevertheless, these environments do not search for semantic state representations, in which an RL algorithm can be better evaluated and controlled. 

More recently, the structure and compositionality of natural language has been used for representing policies in hierarchical RL. In a paper by \citet{hu2019hierarchical}, instructions given in natural language were used in order to break down complex problems into high-level plans and lower-level actions. Their suggested framework leverages the structure inherent to natural language, allowing for transfer to unfamiliar tasks and situations. This use of semantic structure has also been leveraged by \citet{tennenholtz2019natural}, where abstract actions (not necessarily words) were recognized as symbols of a natural and expressive language, improving performance and transfer of RL agents. 

\section{Discussion and Future Work}
\label{sec: disucssion and future work}

Our results indicate that natural language can outperform, and sometime even replace, vision-based representations. Nevertheless, natural language representations can also have disadvantages in various scenarios. For one, they require the designer to be able to describe the state exactly, whether by a rule-based or learned parser. Second, they abstract notions of the state space that the designer may not realize are necessary for solving the problem. As such, semantic representations should be carefully chosen, similar to the process of reward shaping or choosing a training algorithm. Here, we enumerate three instances in which we believe natural language representations are beneficial: \\

    \noindent\textbf{Natural use-case:} Information contained in both generic and task-specific textual corpora may be highly valuable for decision making. This case assumes the state can either be easily described using natural language or is already in a natural language state. This includes examples such as user-based domains, in which user profiles and comments are part of the state, and the stock market, in which stocks are described by analysts and other readily available text. 3D physical environments such as VizDoom also fall into this category, as semantic segmentation maps can be easily described using natural language. \\
    \textbf{Subjective information:} Subjectivity refers to aspects used to express opinions, evaluations, and speculations. These may include strategies for a game, the way a doctor feels about her patient, the mood of a driver, and more. While our work did not focus on subjective information, it is a form of information that is hard to represent in non-textual forms, including images or feature vectors.\\
    \textbf{Unstructured information:} In these cases, features might be measured by different units, with an arbitrary position in the state's feature vector, rendering them sensitive to permutations. Such state representations are difficult to process using neural networks. As an example, the medical domain may contain numerous features describing the vitals of a patient. These raw features, when observed by an expert, can be efficiently described using natural language. Moreover, they allow an expert to efficiently add subjective information. \\
    
An orthogonal line of research considers automating the process of image annotation. While this work considered a manually constructed language parser, general language generators can be used to automatically extract meaningful textual explanations. In addition, while we have only considered spatial features of the state, subjective and transient information can be efficiently encoded as well.

An additional future direction focuses on agent accountability, a pillar of Explainable Artificial Intelligence \citep{samek2017explainable}. Transforming states into natural language can be used as a tool to discern competence. An RL system that is accountable can explain why it acts in a certain way. The explanation can be reasoning in words, by example, or in any other means interpretable to the other entities that communicate with it. It is thus valuable to design interpertable agents that can leverage language in states, actions, and time.

We have shown that natural language representations help interpret the state of an agent, improving its overall performance. While this has worked well in the ViZDoom environment, it may not hold for all domains. Designers of RL algorithms should consider searching for a semantic representation that fits their needs. We also note that natural language representations can be used to augment specific parts of the state space. They can also act as a means to inject prior knowledge to the state. While this work only takes a first step toward finding better semantic state representations, we believe the structure inherent in natural language can be considered a favorable candidate for achieving this goal.

\bibliography{ijcai2020.bib}
\bibliographystyle{plainnat}

\end{document}